\documentclass[letterpaper, 10 pt, conference]{ieeeconf}
\IEEEoverridecommandlockouts
\pdfoutput=1
\usepackage{graphicx} %
\usepackage{amsmath} %
\usepackage{amsfonts}  %
\usepackage{subcaption} %
\usepackage{paralist} %
\usepackage[nostyles]{glossaries}
\usepackage{cite}
\usepackage{tikz}
\usepackage{pgfplots}
\usepackage{url}
\usetikzlibrary{quotes,angles,plotmarks,positioning,backgrounds}
\pgfplotsset{compat=newest}
\usepackage{tikzscale}
\usepackage{hyperref}

\hypersetup{
    pdfauthor     = {Tuomas~V{\"a}lim{\"a}ki and Risto~Ritala},
    pdftitle      = {Optimizing Gaze Direction in a Visual Navigation Task}
}

\newglossary[nlg]{notation}{not}{ntn}{}
\glsdisablehyper

\newcommand{\acrentry}[4][]{ %
  \newacronym[description={#3\ifthenelse{\equal{#4}{}}{}{, {#4}}},#1]{#2}{#2}{#3}
  \glsadd{#2}
}
\newcommand{\notentry}[4][]{ %
    \newglossaryentry{not:#2}{%
        type=notation,
        sort={#2},
        name={#3},
        description={#4},
        #1}
    \glsadd{not:#2}
}
\makeglossaries

\acrentry{QR code}{quick response code}{a two-dimensional barcode}
\acrentry[\glslongpluralkey={Markov decision processes}]{MDP}{Markov decision process}{}
\acrentry[\glslongpluralkey={partially observable Markov decision processes}]{POMDP}{partially observable Markov decision process}{}
\acrentry{GMM}{Gaussian mixture model}{}
\acrentry{ROS}{Robot Operating System}{an open source framework for robotics software development}
\acrentry{PBVI}{point-based value iteration}{}
\acrentry{pdf}{probability density function}{}
\acrentry{MI}{mutual information}{}
\acrentry{EKF}{extended Kalman filter}{}
\acrentry{KL}{Kullback-Leibler}{}
\acrentry{SLAM}{simultaneous localization and mapping}{}

\notentry{Dkl}{$D_{KL}(\cdot||\cdot)$}{Kullback-Leibler divergence}
\notentry{tr}{$\text{tr}(\cdot)$}{trace}
\notentry{ln}{$\text{ln}(\cdot)$}{natural logarithm}
\notentry{N}{$\mathcal{N}(\mu,\Sigma)$}{multivariate normal distribution with mean $\mu$ and covariance $\Sigma$}

\DeclareMathOperator*{\argmax}{arg\,max}
\DeclareMathOperator*{\diag}{diag}

\title{\LARGE \bf Optimizing Gaze Direction in a Visual Navigation Task}

\author{Tuomas~V{\"a}lim{\"a}ki and Risto~Ritala%
\thanks{T.~V{\"a}lim{\"a}ki and R.~Ritala are with the Department of Automation
Science and Engineering, Tampere University of Technology, P.O. Box
692, FI-33101, Tampere, Finland. Email: {\tt\small tuomas.s.valimaki@tut.fi}, {\tt\small risto.ritala@tut.fi}}%
}

\begin{document}

\maketitle
\thispagestyle{empty}
\pagestyle{empty}

\begin{abstract}
Navigation in an unknown environment consists of multiple separable subtasks, such as collecting information about the surroundings and navigating to the current goal. In the case of pure visual navigation, all these subtasks need to utilize the same vision system, and therefore a way to optimally control the direction of focus is needed.
We present a case study, where we model the active sensing problem of directing the gaze of a mobile robot with three machine vision cameras as a \gls{POMDP} using a \gls{MI} based reward function. The key aspect of the solution is that the cameras are dynamically used either in monocular or stereo configuration.  The benefits of using the proposed active sensing implementation are demonstrated with simulations and experiments on a real robot.
\end{abstract}

\section{Introduction}\label{sec:intro}
In human vision, the sequential deployment of gaze in multiple directions is a vital part of information gathering~\cite{Johnson2014}. If a person is given a task to follow a set of driving instructions, they seemingly effortlessly navigate to the desired destination, while simultaneously observing their surroundings for possible landmarks or obstacles. Similar active sensing approaches are used in mobile robotics for e.g. sensor selection. However, there is little research done on actively manipulating individual sensors in a vision system to optimize information gathering.

Active sensing in general refers to seeking a policy for determining the optimal sensor configuration at each time instance as a function of information from previous measurements to achieve a goal~\cite{Hero2011}. The problem can crudely be categorized as an information-gathering problem or as a task-achievement problem depending on whether the goal is to gather maximum amount of information or merely the completion of a task not related to sensing itself. We focus only on the former.

Robots face uncertainty both in predicting and sensing their own state and that of the surroundings. Regardless of whether the goal is to gather information or achieve a task, the active sensing problem becomes a sequential decision making problem in a stochastic environment and can therefore be modeled as a \glsentryfull{POMDP}~\cite{Kaelbling1998}. The utility of actions is measured by a reward function and the robot acts so as to maximize the expected reward. Because the state is not directly observable, the robot's knowledge of it is described as a \gls{pdf} over the state known as the belief state.

Typical information-gathering problems in robotics are exploration tasks. Controlling the direction of sensor focus independently of the robot pose is beneficial when navigating in unknown environments, especially on car-like robots that cannot turn on spot, to gather information about the surroundings. Nevertheless, often only the pose of the robot is considered, and the sensor placements are fixed. In~\cite{Stachniss2005} a solution for integrated \gls{SLAM} and exploration is proposed, using a Rao-Blackwellized particle filter to compute the expected information gain of actions. The approach can be regarded as an approximation of a myopic \gls{POMDP}.

Often the active sensing problems are modeled as task-achievement problems, where the reward function depends only on state and actions and is therefore linear in the belief state and can be solved using standard \gls{POMDP} solvers. Such solutions have been presented e.g. in~\cite{Spaan2009,Spaan2014}. In pure information-gathering problems it is sometimes convenient to use information theoretic rewards that are nonlinear in the belief state. For example~\cite{Araya2010} mentions several rewards typically used for maximizing information, such as the \gls{KL} divergence, which has also been shown to explain human visual attention~\cite{Itti2009}. Other nonstandard \glspl{POMDP} have also been proposed e.g.~\cite{Krishnamurthy2007,Martinez2009}.

\begin{figure}[tb]
  \centering
  \begin{subfigure}{.16\linewidth}
    \includegraphics[width=\linewidth]{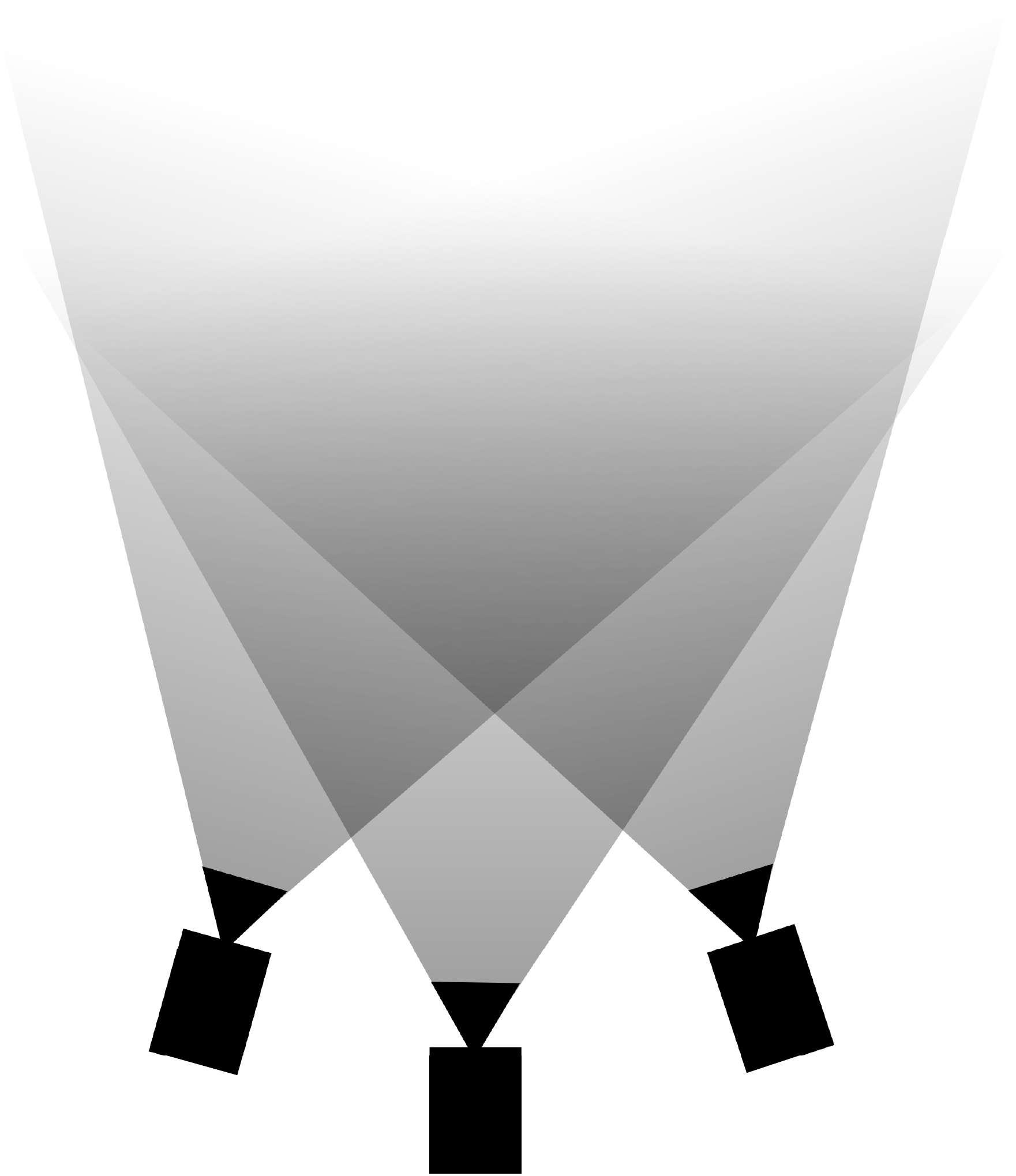}
    \caption{}
  \end{subfigure}
  \begin{subfigure}{.25\linewidth}
    \includegraphics[width=\linewidth]{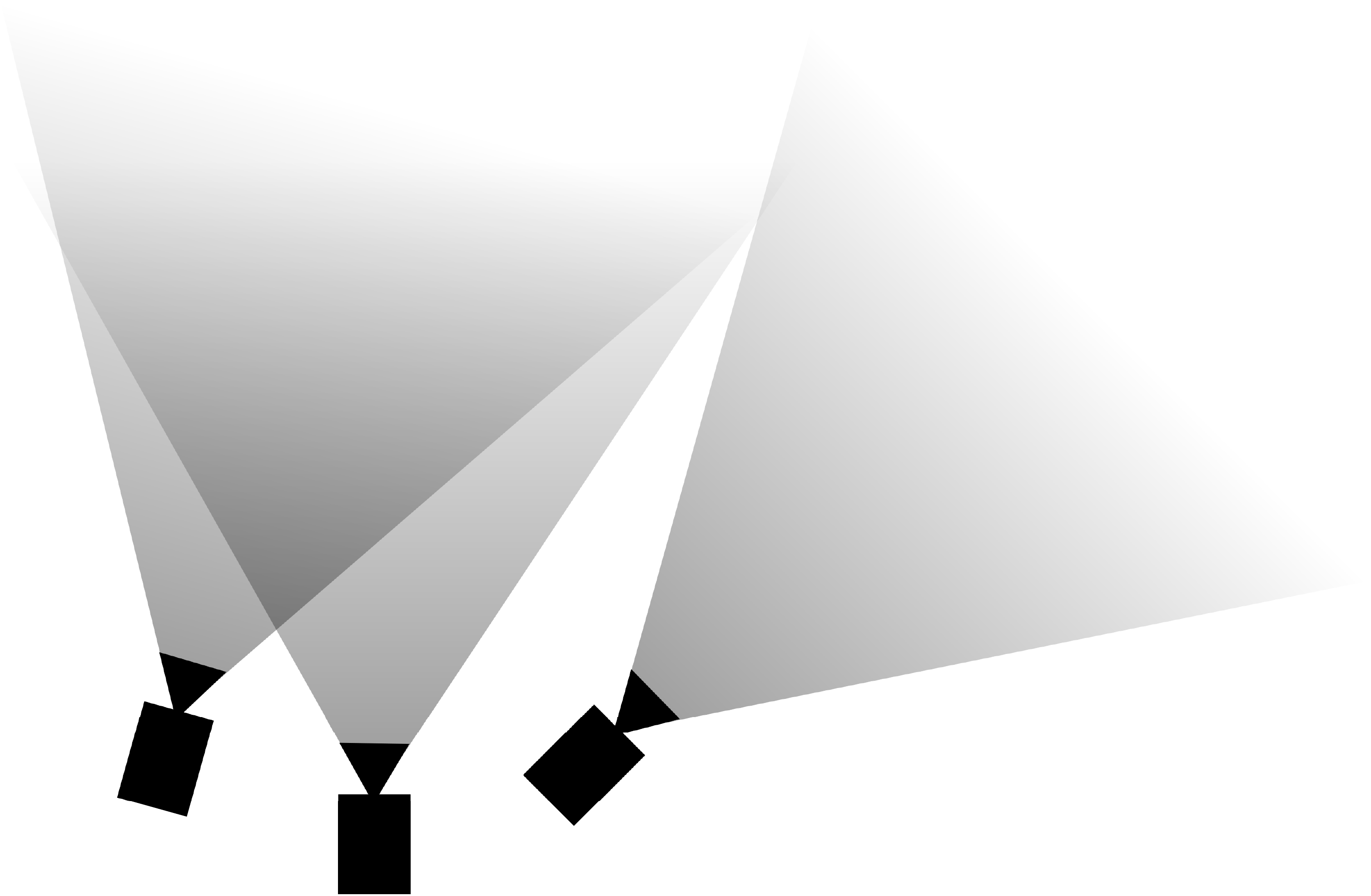}
    \caption{}
  \end{subfigure}
  \begin{subfigure}{.25\linewidth}
    \includegraphics[width=\linewidth]{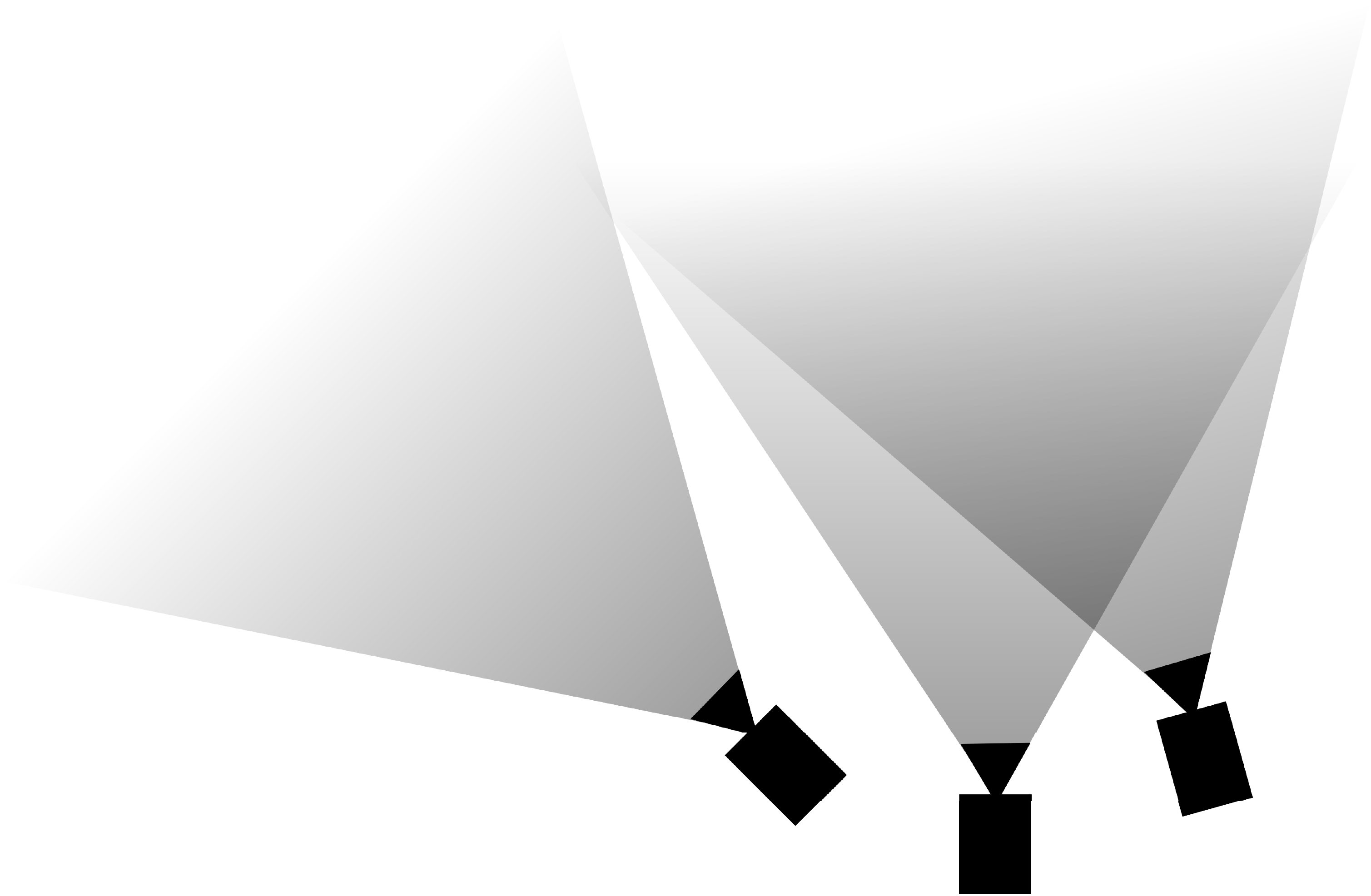}
    \caption{}
  \end{subfigure}
  \begin{subfigure}{.25\linewidth}
    \includegraphics[width=\linewidth]{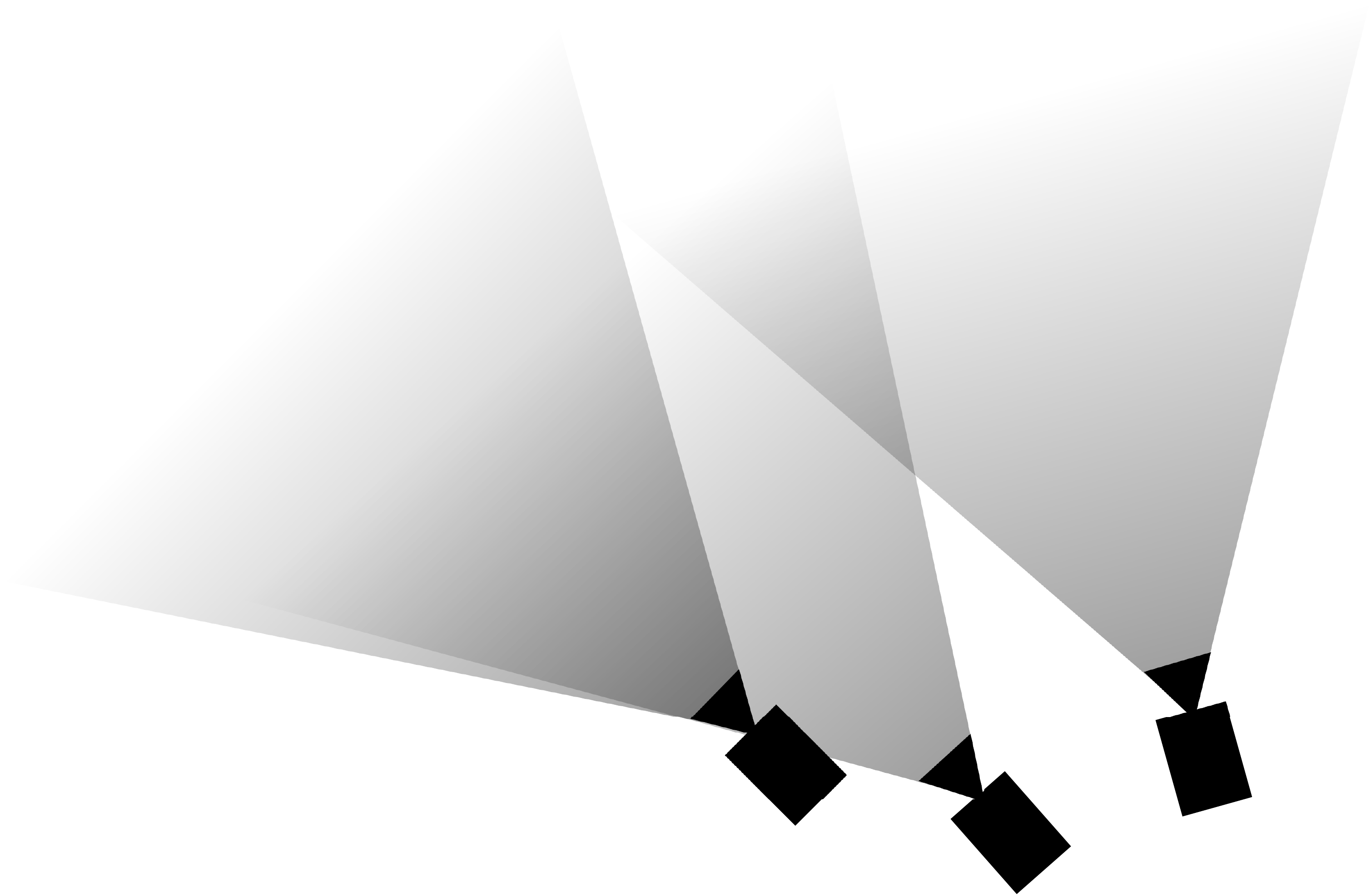}
    \caption{}
  \end{subfigure}
  \begin{subfigure}{.25\linewidth}
    \includegraphics[width=\linewidth]{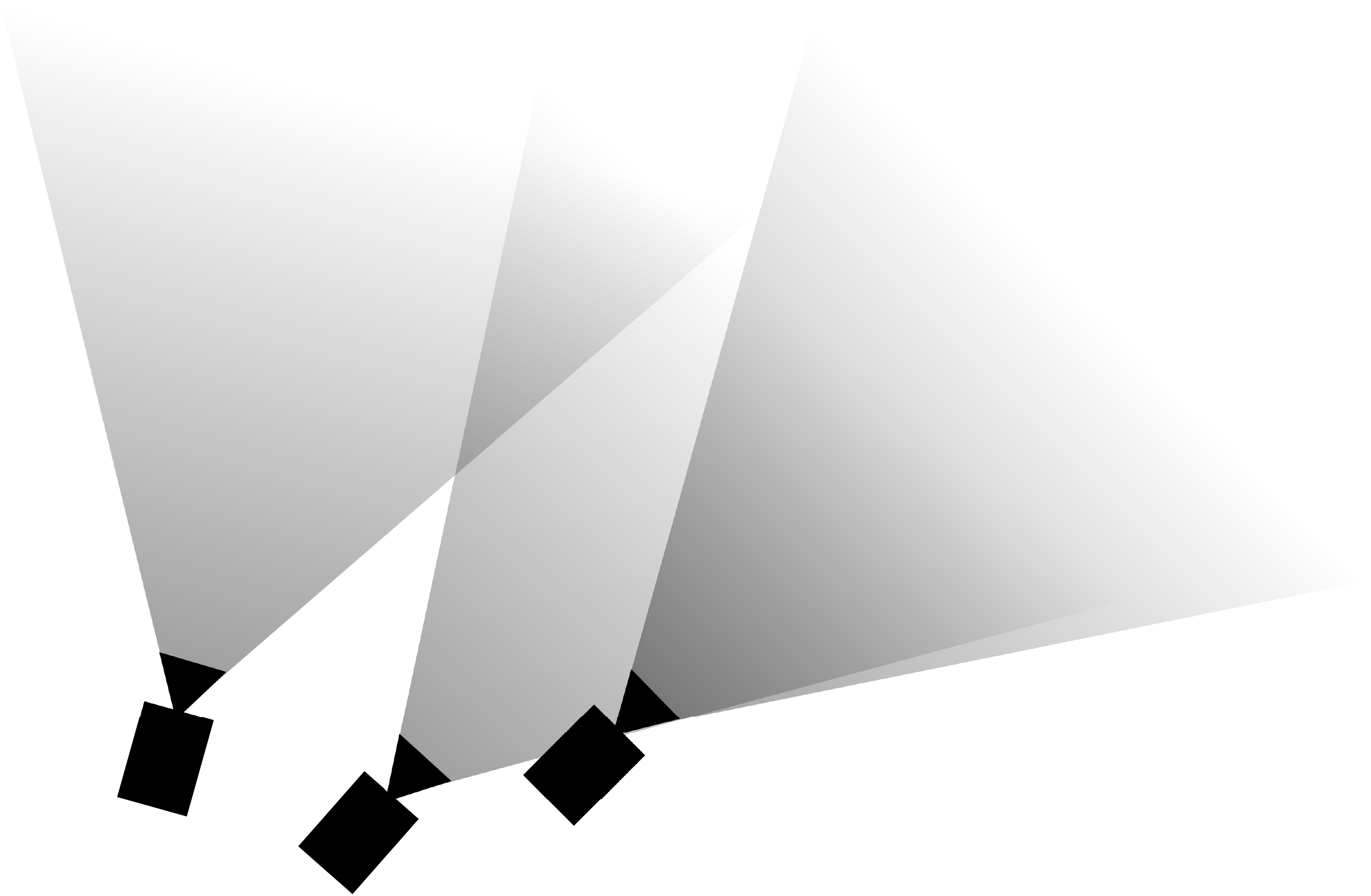}
    \caption{}
  \end{subfigure}
  \begin{subfigure}{.35\linewidth}
    \includegraphics[width=\linewidth]{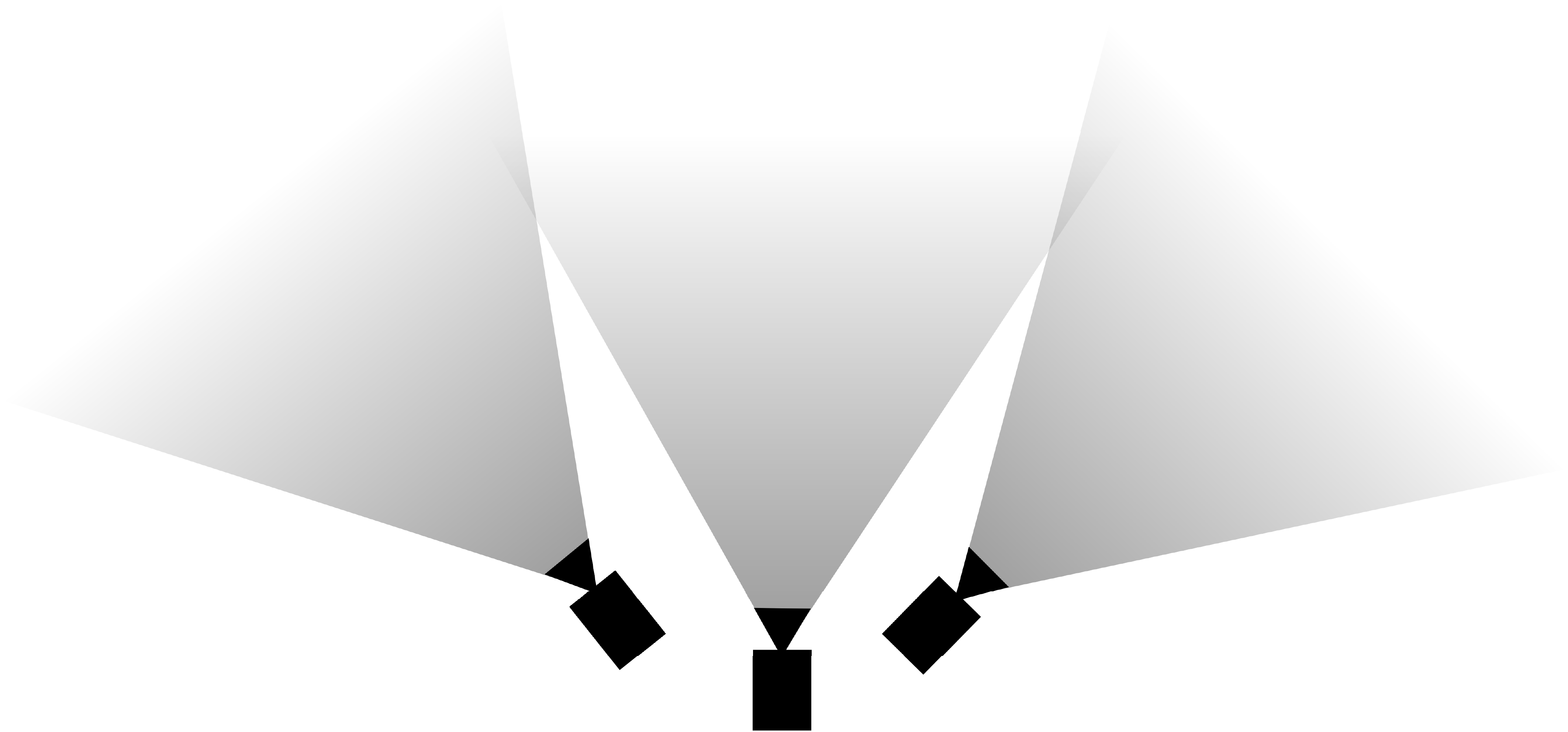}
    \caption{}
  \end{subfigure}
  \caption{The 6 camera modes of the proposed system.}
  \label{fig:cameramodes}
\end{figure}

Recently~\cite{Lauri2014} discusses an active sensing problem, where a robot must decide the direction of focus of its vision system while moving along a fixed trajectory, and proposes a solution via an approximate open-loop feedback control. We study a similar case where the trajectory is not fixed, but consists of landmarks at uncertain locations motivated by driving instructions for human operators; drive straight until you reach a yellow house, turn left and continue a few kilometers until reaching a blue villa.

Our robot must select the direction of gaze of its vision system while navigating a sequence of visual landmarks in order to maximize the amount of information about the positions of the current and next landmark in sequence. As opposed to curiosity driven exploration (e.g.~\cite{Martinez2009}), the landmarks represent navigational goals for an underlying path planner. We formulate the problem as a myopic \gls{POMDP} with information-theoretic reward and solve it online.

The vision system consists of three cameras. The cameras can be in 6 discrete modes as seen in Fig.~\ref{fig:cameramodes} to either maximize the field of view or to focus two or even three cameras in the same direction. When the fields of view of any two cameras overlap, they are used as a stereo pair to allow depth computation as opposed to a monocular camera, where it is only possible to measure the relative angle of the landmark seen in the image w.r.t. the camera. Using the cameras in a monocular configuration is however beneficial when searching for a landmark to narrow down its possible location, which can then be refined by using a stereo pair. The main contribution of this paper is to demonstrate the benefits of using the cameras in a dynamic monocular or stereo configurations depending on the expected information gain, as opposed to a fixed stereo configuration.

The remainder of the paper is organized as follows. In Section~\ref{sec:preliminaries} we outline the basic principles of \glsentrylongpl{POMDP}. Section~\ref{sec:formulation} formulates the case problem and presents the solution. The results of the experiments are reported in Section~\ref{sec:experiments}, and Section~\ref{sec:conclusion} concludes the paper.

\section{Preliminaries}\label{sec:preliminaries}
\Glsentryfullpl{POMDP} are formally presented as the tuple $\langle S,A,T,R,Z,O \rangle$. Consider a sequential decision process, where $s \in S$ denotes the state of the system at any time $t$. At each time step the agent takes an action $a \in A$ and receives a reward $R:S \times A \rightarrow \mathbb{R}$, where $R(s,a)$ is the expected reward of executing action $a$ in state $s$. $T:S \times A \times S \rightarrow [0,1]$ is the state transition model, where $T(s,a,s')=p(s'|s,a)$ describes the probability of ending in state $s'$ when taking the action $a$ in state $s$. In this model, the next state and the expected reward depend only on the current state and the action taken, which is called the \emph{Markov} property. This type of decision processes are called \glspl{MDP}~\cite{Bellman1957,Puterman1994,White1993}.

In case the observations about the state are imperfect and noisy, we model the problem as a \gls{POMDP}~\cite{Smallwood1973,Ross2008}. Let us define $Z$ as the set of all possible observations and $O:S \times A \times Z\rightarrow [0,1]$ as the observation function, where $O(s',a,z')=p(z'|a,s')$ is the probability of observing $z'$ if action $a$ is performed and the resulting state is $s'$.

Because the states are not directly observable, the agent cannot choose its actions based on the state only. Instead it has to consider the whole history of actions and states $h_t = \{a_0,z_1,\dots,a_{t-1},z_t\}$. Maintaining all this information would be memory expensive, but the belief state $b(s)=p(s_t=s|h_t,b_0)$, where $b_0$ is the initial belief about the starting state, is a sufficient statistic for the history~\cite{Smallwood1973}.

The next belief state $b'$ can be calculated from the previous belief state $b$, previous action $a$, and current observation $z'$ using the belief update function $\tau(b,a,z')$ defined as
\begin{align}
  b'(s') &= \tau(b,a,z')(s')\nonumber\\ 
          &=\frac{1}{p(z'|b,a)}O(s',a,z')\sum_{s \in S}T(s,a,s')b(s),
\end{align}
where $p(z'|b,a)$ is the prior probability of observing $z'$ and acts as the normalizing term.

The objective of \gls{POMDP} planning is to find actions that maximize the obtained reward over a time horizon and results in a policy that maps all belief states $b \in B$ to actions; $\pi:B \rightarrow A$. Instead of defining a reward that depends on the state, we define an information theoretic reward $R_b(b,a)$ that depends directly on the belief state. See e.g.~\cite{Araya2010}. As we consider only solutions over one time step i.e. act greedily w.r.t. the expected immediate reward, the optimal policy is
\begin{equation}
  \pi^*=\argmax_\pi\mathbb{E}\left[R_b(b,\pi(b))\right],
\end{equation}
where $R_b(b,\pi(b))$ is the immediate reward for executing policy $\pi$ in belief state $b$.

\section{Problem Formulation}\label{sec:formulation}
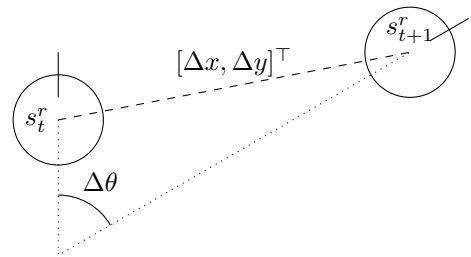
\begin{figure}[tb]
  \medskip
  \centering
  \begin{tikzpicture}[scale = 0.6]
    \coordinate (o) at (0,0);
    \draw ([shift=(90:3)] o) coordinate(a) circle (1) node[left] {$s^r_t$};
    \draw ([shift=(90:0.5)] a) -- ([shift=(90:1.5)] a);
    \draw ([shift=(30:9)] o) coordinate(b) circle (1) node[above] {$s^r_{t+1}$};
    \draw ([shift=(30:0.5)] b) -- ([shift=(30:1.5)] b);
    \draw[dashed] (a) -- (b) node[midway,above] {$[\Delta x,\Delta y]^\top$};
    \draw[dotted] (a) -- (o) -- (b)
      pic["$\Delta\theta$", draw=black,solid,angle eccentricity=1.4,angle radius=0.8cm]
      {angle=b--o--a};
  \end{tikzpicture}
  \caption{Illustration of robot movement between time instances $t$ and $t+1$.}
  \label{fig:transitionmodel}
\end{figure}
\subsection{Case description}
The robot has to select the direction of gaze of its vision system while following a given sequence of landmarks. The \emph{a priori} information of each landmark position is a highly uncertain multivariate Gaussian distribution. The task can be divided into two separate subtasks:
\begin{inparaenum}[1)]
  \item maximizing information about the position of the current landmark in sequence, and
  \item maximizing information about the position of the next landmark in sequence.
\end{inparaenum}
The robot moves at a constant velocity and both subtasks are solved to maximize the probability that the robot actually visits all of the landmarks. When the robot believes it is close enough to the current landmark, it starts navigating to the next.

The three cameras on the robot can be utilized, as shown in Fig.~\ref{fig:cameramodes}, so that either
\begin{inparaenum}[(a)]
  \item all cameras face forward,
  \item left and middle cameras face forward while the right camera faces right,
  \item right and middle cameras face forward while the left camera faces left,
  \item left and middle cameras face left while the right camera faces forward,
  \item right and middle cameras face right while the left camera faces forward, or
  \item all cameras face different directions.
\end{inparaenum}
When the fields of view of any two cameras overlap, they are used as a stereo pair.

\subsection{Dynamics and observation models}\label{sec:models}
The robot's state $s^r_t$ at time instant $t$ is given by its location $x_t$, $y_t$, and its heading $\theta_t$. The robot is controlled by its translational velocity $v_t$, and its rotational velocity $\omega_t$. The robots relative movement between time instances $t$ and $t+1$, as depicted in Fig.~\ref{fig:transitionmodel}, is calculated from the control signals as follows
\begin{equation}
  \begin{bmatrix}
    \Delta x\\\Delta y\\\Delta\theta
  \end{bmatrix} =
  \begin{bmatrix}
    v_t \Delta t \cos(\omega_t\Delta t)\\
    -v_t \Delta t \sin(\omega_t\Delta t)\\
    \omega_t\Delta t
  \end{bmatrix}.
\end{equation}
The control is affected by additive Gaussian noise, so that the joint \gls{pdf} of $u_t=[v_t,\omega_t]^\top$ is $\mathcal{N}(\hat{u_t},Q)$, where $\hat{u_t}$ are the desired control inputs and $Q=\diag(\sigma^2_v,\sigma^2_\omega)$ is the noise covariance.

The robot is set to move forward with a constant translational velocity, $v_t=v$, until it believes to be close enough to the last landmark in sequence. The rotational velocity for each time step, $\omega_t$, is calculated with a discrete time PID controller that minimizes the perpendicular distance of the robot from the desired path---a sequence of linear segments between the landmarks.

The $N$ stationary landmarks are positioned at locations $l^i=\{l^x_i,l^y_i\}^{N}_{i=1}$ and the system state is defined as the coordinates of each landmark in the robot's coordinate frame, $s_t=[l^1_t,l^2_t,\dots,l^N_t]^\top$. The transition model $l^i_{t+1}=f(l^i_t,u_t)$ for each landmark can be written as
\begin{equation}
  l^i_{t+1} = \begin{bmatrix}
                \cos(\Delta\theta) & \sin(\Delta\theta) \\
                -\sin(\Delta\theta) & \cos(\Delta\theta)
              \end{bmatrix} l^i_t - 
              \begin{bmatrix}
                \Delta x \\
                \Delta y
              \end{bmatrix}.
\end{equation}

The robot can observe landmarks that reside inside any of the cameras' cone of observation $C_j(a_t,\alpha,r_{max})_{j=1:3}$ defined by the action $a_t \in A$, view angle $\alpha$, and maximum range $r_{max}$. The action space $A$ is the six camera modes presented in Fig.~\ref{fig:cameramodes}. The camera mode determines which cameras are operated in stereo and which in monocular mode.

In stereo mode the observation consists of the measured landmark coordinates $l^i_t$ and the measurement model $z_t=h_{stereo}(l^i_t)+w_t^{stereo}$ can be written
\begin{equation}
  z_t = l^i_t + w^{stereo}_t,
\end{equation}
where $w^{stereo}_t$ is Gaussian noise with zero mean and covariance $W^{stereo}=\diag(\sigma^2_x,\sigma^2_y)$.
In monocular mode we can only measure the angle $\beta$ of the landmark relative to the robot's coordinate frame and the measurement model $z_t=h_{mono}(l^i_t)+w_t^{mono}$ is
\begin{equation}
  z_t = \beta_t + w^{mono}_t = \arctan\left(\frac{l^{i,y}_t}{l^{i,x}_t}\right) + w^{mono}_t,
\end{equation}
where $w^{mono}_t$ is Gaussian noise with zero mean and variance $W^{mono}=\sigma^2_\beta$.

\subsection{State estimation}\label{sec:estimation}
We apply an \gls{EKF} to track the robot's belief about the landmark locations, $L^i_t|_{i=1:N}$, which are assumed to be independent random variables. Given the previous belief $L^i_{t-1} \sim b^i_{t-1}=\mathcal{N}(\mu^i_{t-1},\Sigma^i_{t-1})$, at each timestep we first calculate the prediction on the following time step $L^{i+}_t \sim b^{i+}_t=\mathcal{N}(\mu^{i+}_t,\Sigma^{i+}_t)$ according to
\begin{subequations}
\begin{align}
  \mu^{i+}_t &= f(\mu^i_{t-1},u_{t-1})\\
  \Sigma^{i+}_t &= F_s \Sigma^i_{t-1} F_s^\top + F_u Q F_u^\top,
\end{align}
\end{subequations}
where $F_s$ and $F_u$ are Jacobians of the transition function $f$ w.r.t. state and control respectfully, evaluated at $(\mu^i_{t-1},u_{t-1})$.

Every time a measurement is received, the posterior $L^i_t|a_{t-1} \sim b^i_t=\mathcal{N}(\mu^i_t,\Sigma^i_t)$ is calculated according to
\begin{subequations}\label{eq:ekfupdate}
\begin{align}
  d_t &= z_t - h(\mu^{i+}_t)\\
  E_t &= H_s \Sigma^{i+}_t H_s^\top + W\\
  K_t &= \Sigma^{i+}_t H_s^\top E_t^{-1}\\
  \mu^i_t &= \mu^{i+}_t + K_t d_t\\
  \Sigma^i_t &= \Sigma^{i+}_t - K_t E_t K_t^\top,
\end{align}
\end{subequations}
where $H_s$ is the Jacobian of the measurement function $h$ w.r.t. state. Depending which measurement is received, the appropriate measurement function $h_{stereo}$ or $h_{mono}$ is used.

\subsection{Reward function}\label{sec:reward}
As the task consist of gaining information about the current and next landmarks in sequence, we need a way to measure the expected information gain. Information theoretic quantities such as the \glsentryfull{KL} divergence~\cite{Kullback1968} are often used to measure the information gain between two \glspl{pdf}. The expected \gls{KL} divergence can be expressed as the \glsentryfull{MI} of the random variables.

We want to maximize the \gls{MI}, $\mathcal{I}(L_{t+1}^{i+},L^i_{t+1}|a_t)$, of the predicted and expected landmark locations, so the total immediate reward is
\begin{align}
  R(b_t,a_t)&=\gamma\mathcal{I}(L_{t+1}^{i+},L^i_{t+1}|a_t)\nonumber\\
            &+\xi\mathcal{I}(L_{t+1}^{i+1+},L^{i+1}_{t+1}|a_t)-g(a_t),
\end{align}
where $\gamma$ and $\xi$ are weighing factors to prioritize gathering information either about the current or about the next landmark and $g(a_t)$ is a cost for changing camera modes defined as
\begin{equation}
  g(a_t)= \begin{cases}
            \zeta, & a_t \neq a_{t-1} \\
            0,     & a_t = a_{t-1},~t=0
          \end{cases},
\end{equation}
where $\zeta$ is a positive scalar constant.

\Gls{MI} of the predictive and expected landmark locations can be calculated as
\begin{align}\label{eq:mi}
  \mathcal{I}(L_{t+1}^{i+},L^i_{t+1}|a_t)
          &= \mathbb{E}_Z\left[D_{KL}(b^{i+}_{t+1},b^i_{t+1})\right]\\
          &= \int_Z D_{KL}(b^{i+}_{t+1},b^i_{t+1})p(z_{t+1}|a_t,b^i_t) \text{d}z_{t+1}\nonumber
\end{align}
where $b^i_{t+1}$ depends on the observation $z_{t+1}$, and $p(z_{t+1}|a_t,b^i_t)$ is the probability that the landmark $l^i_{t+1}$ is visible in observation $z_{t+1}$ if action $a_t$ is taken in current belief $b^i_t$. $D_{KL}(\cdot)$ is the \gls{KL} divergence (see e.g.~\cite{Cover2006}).

The probability of $l^i_{t+1}$ being visible in observation $z_{t+1}$ would be difficult to calculate, so instead we use Monte Carlo methods and draw $K$ samples from the probability distribution; $\{z_{t+1}^k\}_{k=1}^K \sim p(z_{t+1}|a_t,b^i_t)=\mathcal{N}(h(\mu^{i+}_{t+1}),E_{t+1})$. In other words, we sample possible landmark locations from $\mathcal{N}(\mu^{i+}_{t+1},\Sigma^{i+}_{t+1})$ and expect to get a measurement $z_{t+1}^k$ from each sample that lies inside the appropriate cones of observation $C_j(a_t,\alpha,r_{max})_{j=1:3}$. The camera mode $a_t$ determines which cameras are used as a stereo pair, i.e. the sample must be inside both cones of observation, and which as monocular cameras. Now (\ref{eq:mi}) can be approximated with
\begin{equation}
  \mathcal{I}(L_{t+1}^{i+},L^i_{t+1}|a_t) \approx \frac{1}{K}\sum_{k=1}^K D_{KL}(b^{i+}_{t+1},b^i_{t+1}),
\end{equation}
where $b^i_{t+1}$ depends on the sampled measurement $z_{t+1}^k$.

\subsection{Solving the problem}\label{sec:solving}
Acting greedily may result in poorer performance if
\begin{inparaenum}[1)]
  \item the focus of attention cannot be changed rapidly, or
  \item the observed features are not stationary
\end{inparaenum}~\cite{Lauri2014}.
As these do not apply to the problem defined above, we consider only myopic optimization of the immediate information gain. 

At timestep $t$ we solve
\begin{equation}
  a_t^* = \argmax_{a_t \in A}R(b_t,a_t)
\end{equation}
to obtain the optimal action.

\section{Experiments}\label{sec:experiments}
\begin{figure}[!ht]
  \centering
  \medskip
  \begin{subfigure}{0.95\linewidth}%
    \centering
    \input{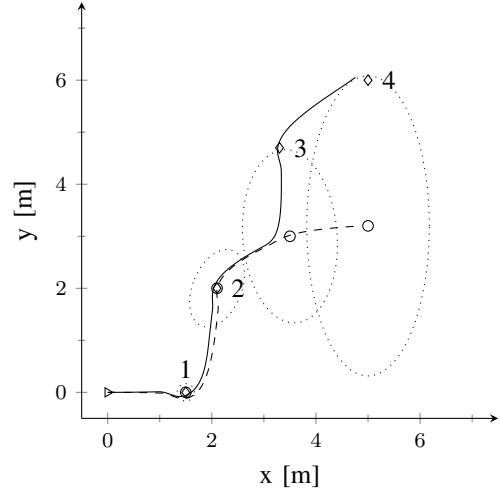}
    \caption{With fixed front facing stereo cameras the robot fails to see the third landmark and continues towards the prior mean position. When operated with the active sensing enabled the robot observes the actual landmark and corrects its trajectory accordingly.}
    \label{fig:simulation}
  \end{subfigure}
  \begin{subfigure}{0.95\linewidth}%
    \bigskip
    \centering
    \input{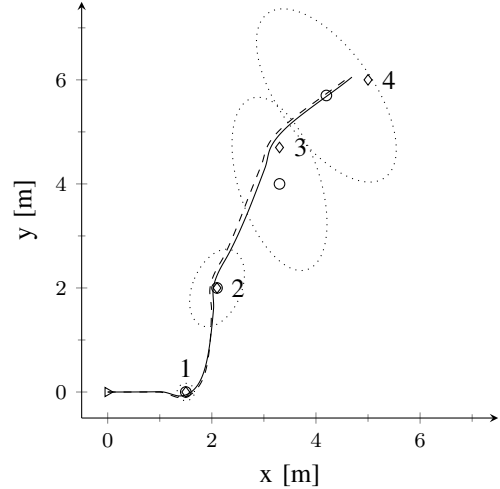}
    \caption{When the prior means are close to the actual landmark positions, there is little difference in the performance of the two methods tested.}
    \label{fig:simulation2}
  \end{subfigure}
  \caption{Simulation experiments both with and without using the active sensing. The initial position of the robot is marked with the triangle. Circular markers denote the prior of the landmarks, with the associated dotted ellipses representing the 50\% confidence intervals. Diamond markers denote the actual positions of the landmarks. The dashed line is the robot trajectory when active sensing is not used and the solid line is the robot trajectory when using the proposed active sensing.}
  \label{fig:simulations}
\end{figure}

\subsection{Simulation experiments}\label{sec:simulation}
We demonstrate with simulation experiments the benefits of using the above described active sensing implementation with dynamic camera configurations as opposed to just using fixed front-facing stereo cameras, i.e. the cameras are permanently in camera mode (a) of Fig.~\ref{fig:cameramodes}. Both methods were tested in two scenarios with different prior information. The experiment was repeated 10 times for each method and scenario.

The sequence of landmarks in the simulation environment consisted of $N=4$ landmarks. Different initial beliefs of the landmark positions with 50\% confidence intervals can be seen in Fig.~\ref{fig:simulations} with the actual landmark positions, and the initial position of the robot. The gaze direction of the cameras is $40^\circ$ when looking left, $-40^\circ$ when looking right, and $0^\circ$ when facing forward. The angle of view of the cameras is $\alpha=50^\circ$ and maximum range $r_{max}=2.5$~m. The vision system outputs observations at 1 Hz rate. The noise parameters used were $\sigma^2_v=\sigma^2_\omega=0.01~\text{m}^2/\text{s}^2$ and $\text{rad}^2/\text{s}^2$ accordingly, $\sigma^2_x=0.01~\text{m}^2$, $\sigma^2_y=0.02~\text{m}^2$, and $\sigma^2_\beta=0.05~\text{rad}^2$. The number of samples to draw for the Monte Carlo sampling was $K=10$. The weighing factors were $\gamma=0.7$, $\xi=0.3$, and the cost $\zeta=0.01$. The robot starts navigating to the next landmark when it believes to be closer than 0.2 m to the current landmark in sequence.

\begin{figure}[t]
  \centering
  \smallskip
  \begin{subfigure}{0.82\linewidth}
    \includegraphics[width=\linewidth,axisratio=1.5]{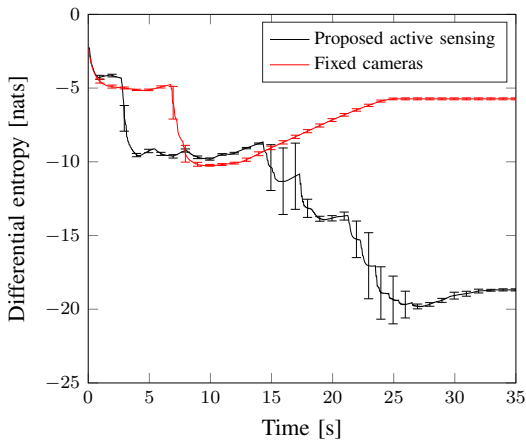}
    \caption{Prior distribution as depicted in Fig.~\ref{fig:simulation}.}
    \label{fig:entropy}
  \end{subfigure}
  \begin{subfigure}{0.82\linewidth}
  \medskip
    \includegraphics[width=\linewidth]{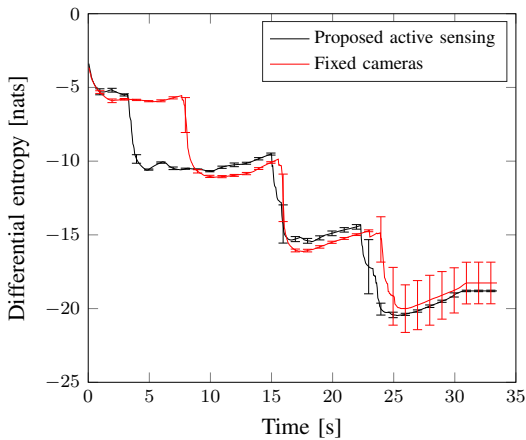}
    \caption{Prior distribution as depicted in Fig.~\ref{fig:simulation2}.}
    \label{fig:entropy2}
  \end{subfigure}
  \caption{Time evolution of belief state entropy. The lines indicate mean differential entropy over 10 experiments and the bars indicate 95\% confidence intervals.}
  \label{fig:entropies}
\end{figure}
Fig.~\ref{fig:simulations} presents typical outcomes of the simulations in both tested scenarios. In the case of Fig.~\ref{fig:simulation} the robot with fixed cameras never reaches all of the landmarks because it fails to see the actual third landmark situated to the left of its originally planned trajectory. When the proposed active sensing is used instead, the robot keeps observing its surroundings in order to minimize the uncertainty about the landmark and upon observing the actual landmark, corrects its belief and plans a new trajectory to follow.

To quantify the results, the differential entropy of the belief state during the experiments was studied and can be seen in Fig.~\ref{fig:entropy}. Lower entropy indicates lower uncertainty. Using the proposed active sensing, the robot visited all the landmarks on each experiment, whereas the robot with fixed cameras never observed the third and fourth landmarks and only navigated to the prior mean locations. This can be seen also from the development of differential entropy; after the second sharp decline, where the robot first observes a new landmark, the differential entropy starts increasing due to the control noise, and finally reaches a steady state when the robot stops. With the proposed active sensing the differential entropy keeps declining when new landmarks are observed, and a much lower final value is reached.

If, however, the actual landmark positions are close to the prior means, there is little difference between the performance of the proposed active sensing and fixed cameras as seen in Fig.~\ref{fig:simulation2}. From the sharp declines of differential entropy of the belief state in Fig.~\ref{fig:entropy2}, we can see that the active sensing method intuitively observes the landmarks sooner than the method with fixed front-facing cameras. The fixed camera method, however, reaches lower differential entropy values after observing a landmark, because all three cameras are used as stereo pairs, whereas the active sensing methods usually only observes the landmark with one stereo pair, and operates the third camera in a monocular mode. There is no significant difference in the final entropy of the two methods. The fixed camera method failed to observe the fourth landmark in one of the experiments, which resulted in the large confidence intervals seen at the end. 

Using the proposed active sensing is especially beneficial when the prior information about the landmark positions is highly uncertain and the actual landmark positions differ from the prior means. Random actions in the dynamic configuration could lead to similar results, but would be impractical and wear out a physical system.

\subsection{Experiments on a real robot}\label{sec:realexperiment}
\begin{figure}[t]
  \medskip
  \centering
  \includegraphics[width=.3\linewidth]{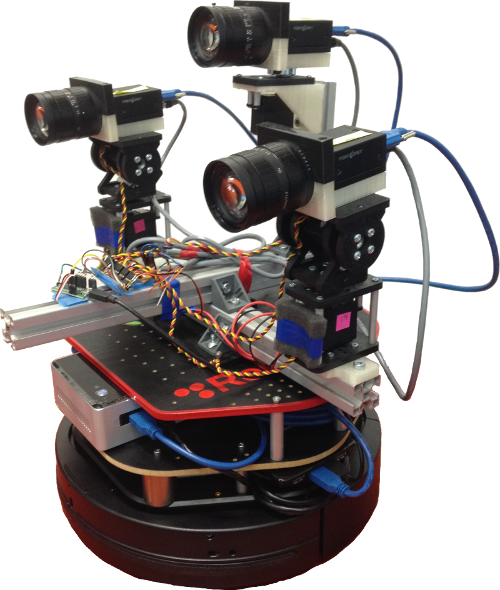}
  \caption{Turtlebot equipped with the camera rig, which was used to carry out the experiments described in Section~\ref{sec:realexperiment}.}
  \label{fig:robot}
\end{figure}

\begin{figure}
  \centering
  \input{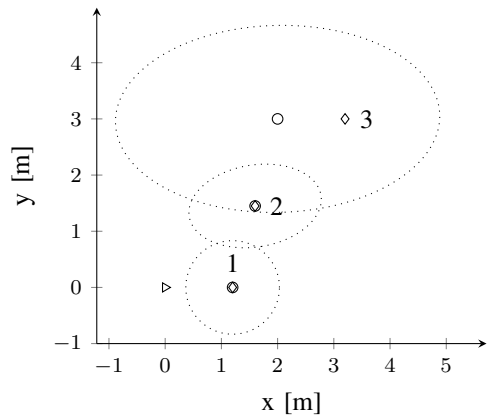}
  \caption{Setting of the real world experiment. Initial position of the robot is marked with the triangle. Circular markers denote the prior of the landmarks, with the associated dotted ellipses representing the 50\% confidence intervals. Diamond markers denote the actual positions of the landmarks.}
  \label{fig:turtlecase}
\end{figure}
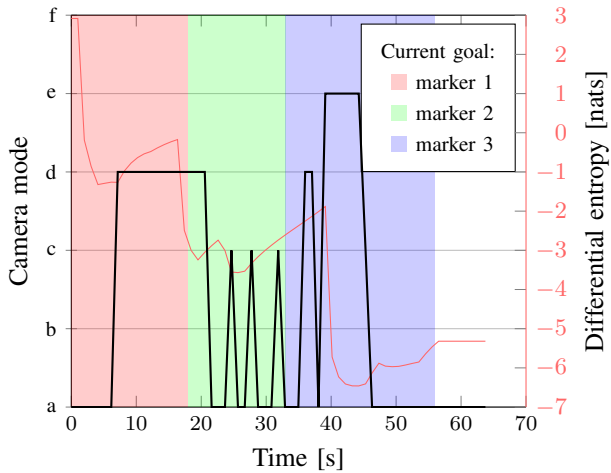
\begin{figure}
  \centering
  \vspace*{.5ex}
  \begin{tikzpicture}
\begin{axis}[small,%
width=0.7\linewidth,
scale only axis,
xmin=0,
xmax=70,
xlabel={Time [s]},
ymin=0,
ymax=5,
ytick={0,1,2,3,4,5},
yticklabels={{a},{b},{c},{d},{e},{f}},
ylabel={Camera mode},
xlabel near ticks,
axis y line=left,
y axis line style=-,
ylabel near ticks,
ymajorgrids
]
\addplot [draw=none,fill=red,fill opacity=0.2]
  coordinates{(0,0) (18,0) (18,5) (0,5)};
\addplot [draw=none,fill=green,fill opacity=0.2]
  coordinates{(18,0) (33,0) (33,5) (18,5)};
\addplot [draw=none,fill=blue,fill opacity=0.2]
  coordinates{(33,0) (56,0) (56,5) (33,5)};

\node at (axis cs:68.5,4.9) [anchor=north east,inner sep=0pt]{
    \begin{tikzpicture}[show background rectangle,
    background rectangle/.style={draw,fill=white},
    node distance=1.2ex,
    every node/.style={font=\footnotesize,inner sep=3pt}
    ]
    \node (title) {Current goal:};
    \node[below=of title.west,xshift=1em,yshift=-.5ex,rectangle,fill=red,fill opacity=0.2,label=east:marker 1] (1) {};
    \node[below=of 1,rectangle,fill=green,fill opacity=0.2,label=east:marker 2] (2) {};
    \node[below=of 2,rectangle,fill=blue,fill opacity=0.2,label=east:marker 3] (3) {};
    \end{tikzpicture}
    };
\end{axis}

\begin{axis}[small,%
width=0.7\linewidth,
scale only axis,
xmin=0,
xmax=70,
every outer y axis line/.append style={red!60},
every y tick label/.append style={font=\small\color{red!60}},
ymin=-7,
ymax=3,
ytick={-7, -6, -5, -4, -3, -2, -1,  0,  1,  2,  3},
ylabel={Differential entropy [nats]},
hide x axis,,
axis y line*=right,
ylabel near ticks,
yticklabel style={text width=1.2em,align=right}
]
\addplot [color=red!60,solid,forget plot]
  table[row sep=crcr]{%
0   2.91596210043989\\
1.00002861022949    2.91596210043989\\
2.00466465950012    -0.187325862922574\\
3.05001449584961    -0.856468433889339\\
4.099933385849  -1.32192159875001\\
5.1025562286377 -1.29108718592315\\
6.15312600135803    -1.26042371610758\\
7.14995050430298    -1.26614501380872\\
8.1999237537384 -0.968231408000599\\
9.2000093460083 -0.773708625131182\\
10.1999809741974    -0.634465702093032\\
11.2000253200531    -0.539290038544002\\
12.2499477863312    -0.478403579185248\\
13.2499194145203    -0.386550640992368\\
14.2999498844147    -0.302602551768119\\
15.3499553203583    -0.230428097870202\\
16.3499460220337    -0.173561289240453\\
17.3999674320221    -2.49217843843874\\
18.4499535560608    -3.00220077853905\\
19.5005416870117    -3.24337141739872\\
20.5499930381775    -3.05038954729396\\
21.5999941825867    -2.88264417452495\\
22.6000499725342    -2.74052712344662\\
23.6545255184174    -3.00464899164571\\
24.6500034332275    -3.55291418585558\\
25.6547634601593    -3.56927876124415\\
26.7010517120361    -3.53238601316721\\
27.7500095367432    -3.3304626793899\\
28.7500522136688    -3.16382560508071\\
29.8001115322113    -3.0075036374405\\
30.80930352211  -2.87147329546317\\
31.85165143013  -2.73805366074997\\
32.9004871845245    -2.6121432104644\\
33.9008374214172    -2.49557426865584\\
34.949934720993 -2.3754362499351\\
35.9999508857727    -2.25517839758898\\
37.0499374866486    -2.1328805662418\\
38.0501480102539    -2.01309107576822\\
39.099942445755 -1.88283812418803\\
40.1499516963959    -5.72195857875973\\
41.1499388217926    -6.22914075346512\\
42.1999344825745    -6.41109827551773\\
43.249885559082 -6.45877575313008\\
44.2499771118164    -6.46028137515519\\
45.2999002933502    -6.40810807951232\\
46.3000180721283    -6.13098305611937\\
47.349898815155 -5.88113974871209\\
48.3499319553375    -5.95535970816644\\
49.3999607563019    -5.96936011771247\\
50.3999395370483    -5.96124356262163\\
51.4499447345734    -5.92665376166203\\
52.5077302455902    -5.88566330126959\\
53.5019521713257    -5.85041184649987\\
54.5500037670135    -5.63958950171307\\
55.5500504970551    -5.46327307126965\\
56.5500197410583    -5.31933717741577\\
57.5999805927277    -5.31933717741577\\
58.6548190116882    -5.31933717741577\\
59.6499259471893    -5.31933717741577\\
60.7033579349518    -5.31933717741577\\
61.7000629901886    -5.31933717741577\\
62.7037751674652    -5.31933717741577\\
63.7498791217804    -5.31933717741577\\
};
\end{axis}

\begin{axis}[small,%
width=0.7\linewidth,
scale only axis,
xmin=0,
xmax=70,
ymin=0,
ymax=5,
hide axis
]
\addplot [solid,forget plot,thick]
  table[row sep=crcr]{%
0	0\\
1.00002861022949	0\\
2.00466465950012	0\\
3.05001449584961	0\\
4.099933385849	0\\
5.1025562286377	0\\
6.15312600135803	0\\
7.14995050430298	3\\
8.1999237537384	3\\
9.2000093460083	3\\
10.1999809741974	3\\
11.2000253200531	3\\
12.2499477863312	3\\
13.2499194145203	3\\
14.2999498844147	3\\
15.3499553203583	3\\
16.3499460220337	3\\
17.3999674320221	3\\
18.4499535560608	3\\
19.5005416870117	3\\
20.5499930381775	3\\
21.5999941825867	0\\
22.6000499725342	0\\
23.6545255184174	0\\
24.6500034332275	2\\
25.6547634601593	0\\
26.7010517120361	0\\
27.7500095367432	2\\
28.7500522136688	0\\
29.8001115322113	0\\
30.80930352211	0\\
31.85165143013	2\\
32.9004871845245	0\\
33.9008374214172	0\\
34.949934720993	0\\
35.9999508857727	3\\
37.0499374866486	3\\
38.0501480102539	0\\
39.099942445755	4\\
40.1499516963959	4\\
41.1499388217926	4\\
42.1999344825745	4\\
43.249885559082	4\\
44.2499771118164	4\\
45.2999002933502	2\\
46.3000180721283	0\\
47.349898815155	0\\
48.3499319553375	0\\
49.3999607563019	0\\
50.3999395370483	0\\
51.4499447345734	0\\
52.5077302455902	0\\
53.5019521713257	0\\
54.5500037670135	0\\
55.5500504970551	0\\
56.5500197410583	0\\
57.5999805927277	0\\
58.6548190116882	0\\
59.6499259471893	0\\
60.7033579349518	0\\
61.7000629901886	0\\
62.7037751674652	0\\
63.7498791217804	0\\
};
\end{axis}
\end{tikzpicture}%
  \caption{Time evolution of used camera modes and the belief state entropy.}
  \label{fig:turtlemode}
\end{figure}

The robot seen in Fig.~\ref{fig:robot} was used to demonstrate the developed algorithms on physical hardware. The robot features three Point Grey Grasshopper3 USB3 machine vision cameras with $2048\times2048$ resolution mounted on a Turtlebot platform. The gaze direction of the cameras can be controlled between $[-\frac{\pi}{2},\frac{\pi}{2}]$ radians. The angle of view of the cameras is $\alpha=50^\circ$. QR codes are used as landmarks since they are easy to recognize from the images and identify. Successfully identifying the QR codes limits the maximum range to $r_{max}=4$~m. The on-board computer of the robot is an Intel NUC with a 3.1 GHz i7 processor and 8 GB of RAM.

The initial setting of the experiment is pictured in Fig.~\ref{fig:turtlecase}. The mean of the prior of the third landmark is off by so much, that the landmark is not observed by static front facing cameras. Behavior of the developed active sensing is visible in Fig.~\ref{fig:turtlemode}. At first, uncertainty of first landmark is reduced by directing all cameras towards it. After that, the algorithms prefers to keep one camera directed at the first landmark, and a stereo pair towards the second landmark. Between time-instances 21 and 39, the camera mode is frequently changed to observe the highly uncertain third landmark. After the observation is acquired, the cameras are directed at the landmark. Demonstration of the system in action can be found on \url{https://www.youtube.com/watch?v=n6AWOLpbNzs}.

\section{Conclusion}\label{sec:conclusion}
We studied an active sensing problem where three cameras are operated in six configurations, and used as either monocular cameras or stereo pairs. A robot follows a trajectory defined as a sequence of visual landmarks at uncertain locations and has to direct its gaze to maximize information about the landmark locations. The problem was formulated as a \gls{POMDP} with a reward function based on mutual information and a myopic solution was provided.

Simulation experiments demonstrate the benefits of using the proposed active sensing implementation over using fixed front-facing stereo pairs. Our active sensing solution outperforms the fixed camera solution when prior information is highly uncertain and the prior mean differs from the actual landmark positions. If prior information is accurate or the prior means happen to be close to the actual landmark positions, there is little difference between the two tested methods. There was, however, no disadvantage in using the active sensing in any of the tests performed. Real world experiments demonstrate that the algorithm runs real-time on physical hardware.

The experiments are not comprehensive and can be only consider proof-of-concept, but the method shows promising results that motivate further development. In future work, the solution will be generalized, and we plan to apply similar active sensing methods to a wider range of problems in mobile robotics.

\section*{Acknowledgment}
The authors would like to thank Mikko Lauri for his helpful comments and suggestions. The part by Risto Ritala was funded by the Academy of Finland, project \emph{``Optimal operation of observation subsystems in autonomous mobile machines''}, which is gratefully acknowledged.

\bibliographystyle{IEEEtran}
\bibliography{IEEEabrv,bibliography}

\end{document}